\pdfoutput=1

\documentclass[11pt]{article}

\usepackage{acl}

\usepackage{times}
\usepackage{latexsym}

\usepackage[T1]{fontenc}

\usepackage[utf8]{inputenc}

\usepackage{microtype}
\usepackage[normalem]{ulem}
\usepackage{multirow}
\usepackage{algorithm}
\usepackage{algorithmic}
\usepackage{graphicx}
\usepackage{amsmath}
\usepackage{amsfonts}
\usepackage{bm}
\usepackage{booktabs}
\usepackage{subfigure}
\usepackage{verbatim}
\usepackage{comment}
\usepackage{arydshln}

\usepackage{enumitem}

\usepackage{bbm}
\title{BvSP: Broad-view Soft Prompting for Few-Shot Aspect Sentiment Quad Prediction}

\author{Yinhao Bai\textsuperscript{1} \quad Yalan Xie\textsuperscript{2} \quad Xiaoyi Liu\textsuperscript{2} \quad Yuhua Zhao\textsuperscript{1} \quad {\bf Zhixin Han\textsuperscript{1}} \\ \quad {\bf Mengting Hu\textsuperscript{1}\thanks{\; Corresponding author.}} \quad {\bf Hang Gao\textsuperscript{3}} \quad {\bf Renhong Cheng\textsuperscript{2}} \\
\textsuperscript{1} College of Software, Nankai University, \textsuperscript{2} College of Computer Science, Nankai University, \\ \textsuperscript{3} College of Artificial Intelligence, Tianjin University of Science and Technology \\
{\tt \{yinhao,xyl,xyiliu,zyh22,zhixinhan\}@mail.nankai.edu.cn,} \\ {\tt mthu@nankai.edu.cn}
}

\begin{document}
\maketitle
\begin{abstract}
Aspect sentiment quad prediction (ASQP) aims to predict four aspect-based elements, including \emph{aspect term}, \emph{opinion term}, \emph{aspect category}, and \emph{sentiment polarity}. In practice, unseen aspects, due to distinct data distribution, impose many challenges for a trained neural model. Motivated by this, this work formulates ASQP into the few-shot scenario, which aims for fast adaptation in real applications. Therefore, we first construct a few-shot ASQP dataset ($\mathtt{FSQP}$) that contains richer categories and is more balanced for the few-shot study. Moreover, recent methods extract quads through a generation paradigm, which involves converting the input sentence into a templated target sequence. However, they primarily focus on the utilization of a single template or the consideration of different template orders, thereby overlooking the correlations among various templates. To tackle this issue, we further propose a \emph{Broad-view Soft Prompting} (BvSP) method that aggregates multiple templates with a broader view by taking into account the correlation between the different templates. Specifically, BvSP uses the pre-trained language model to select the most relevant k templates with Jensen–Shannon divergence. BvSP further introduces soft prompts to guide the pre-trained language model using the selected templates. Then, we aggregate the results of multi-templates by voting mechanism. 
 Empirical results demonstrate that BvSP significantly outperforms the state-of-the-art methods under four few-shot settings and other public datasets. Our code and dataset are available at \url{https://github.com/byinhao/BvSP}.
\end{abstract}

\section{Introduction}

Analyzing user reviews, social media posts, product evaluations, and other content on the web to extract sentiment information related to specific aspects helps in understanding users' opinions and emotions regarding different aspects on the web. 
To monitor public opinion and support decision-making, the research field of sentiment analysis and opinion mining emerged \cite{vinodhini2012sentiment,shaik2022sentiment}. The aspect sentiment quad prediction (ASQP) task aims to extract aspect quadruplets from a review sentence to comprehensively understand users' aspect-level opinions \cite{li2023aspect,hu2022improving}. Recently, ASQP is gaining attention due to it involves predicting four fundamental aspect-level elements: 1) \emph{aspect term} which is the concrete aspect description in the given text; 2) \emph{opinion term} describing the exact opinion expression towards an aspect; 3) \emph{aspect category} denoting the aspect type which refers to a pre-defined set, and 4) \emph{sentiment polarity} indicating the sentiment class of the aspect. For example, given the sentence  \emph{``The room is clean.''}, the aspect elements are “room”, “clean”, “room\_overall”, and “positive”, respectively. Accordingly, the ASQP is described as a quad (\emph{room}, \emph{clean}, \emph{room\_overall}, \emph{positive}). 

\begin{figure}[t]
\centering
\includegraphics[width=0.48\textwidth]{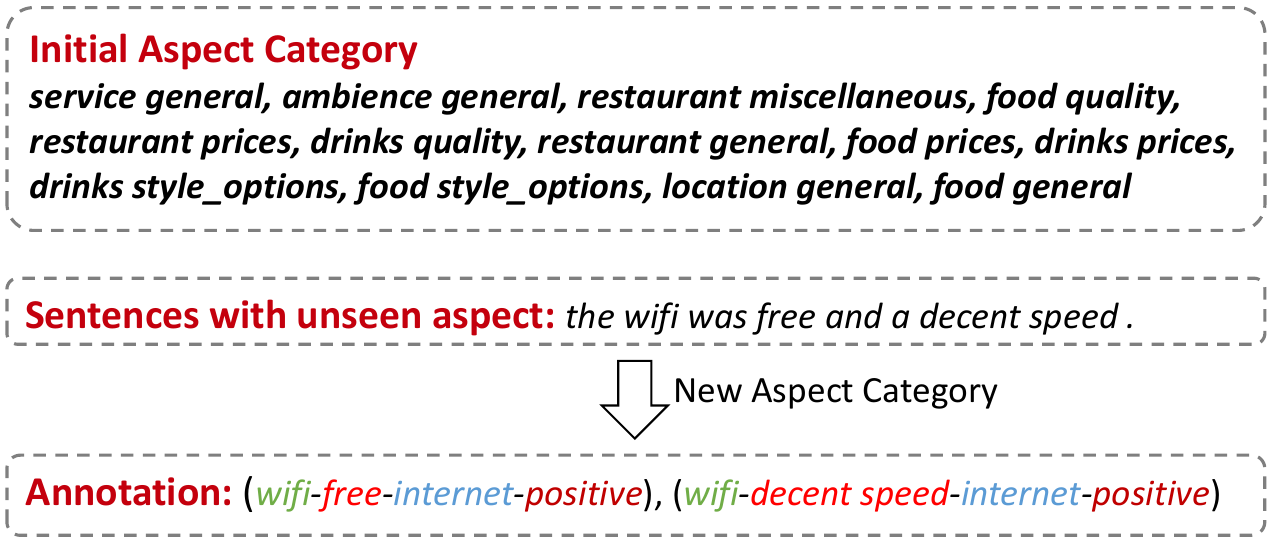} 
\caption{A unseen aspect case is shown. The newly emerged category \emph{`internet''} is not mentioned in the pre-defined set of aspect categories.}
\label{fig:unseen_aspect}
\end{figure}

However, in practical situations, aspect categories are not immutable and frozen \cite{zhou2022semantic}. 
New aspects emerge as people discuss emerging phenomena, trends, products, and more through social media, news articles, and other means on the internet.
As the restaurant domain illustrated in Figure \ref{fig:unseen_aspect}, the initial aspect category set is pre-defined. Yet as the infrastructure upgrades, new aspects, such as \emph{``WiFi''}, gradually appear. The sentence's category, i.e. \emph{``internet''} does not exist in the pre-defined categories. This imposes challenges to the model's comprehensive and accurate understanding of the sentence. Moreover, the unseen aspect usually has a distribution shift, which is struggling for trained models to adapt accurately.

\tabcolsep=0.5cm
\begin{table*}[]
\renewcommand\arraystretch{1.3}
\small
    \centering
    
    \setlength{\tabcolsep}{1.4mm}{
    \begin{tabular}{l|ccc|cccccc|cc}
    \hline
     \multirow{2}{*}{Dataset} & \multicolumn{3}{c|}{Sentence} & \multicolumn{6}{c|}{Quad} & \multicolumn{2}{c}{Category} \\
     & \#S & \#W & \#W/S & EA \& EO & IA \& EO & EA \& IO & IA \& IO & \#Q & \#Q/S & \#C & \#M(C) \\
    \hline
    $\mathtt{Rest15}$ & 1580 & 22886 & 14.48 & 1946 & 550 & - & - & 2496 & 1.57 & 13 & 192 \\
    $\mathtt{Rest16}$ & 2124 & 30805 & 14.50 & 2566 & 729 & - & - & 3295 & 1.55 & 13 & 253 \\
    $\mathtt{Restaurant}$ & 2284 & 34417 & 15.06 & 2431 & 530 & 350 & 350 & 3661 & 1.60 & 13 & 281 \\
    $\mathtt{Laptop}$ & 4076 & 63879 & 15.67 & 3278 & 912 & 1241 & 342 & 5773 & 1.41 & 121 & 47 \\
    \hline
    $\mathtt{FSQP}$ (Ours) & 12551 & 149016 & 11.87 & 10749 & 151 & 5185 & 298 & 16383 & 1.31 & 80 & 205 \\
    
    \hline
    \end{tabular}}
    \caption{Data statistics and comparisons. 
    \#S, \#W, \#Q, and \#C denote the number of sentences, words, quads, and categories, respectively. EA, EO, IA, and IO denote explicit aspect, explicit opinion, implicit aspect, and implicit opinion. \#M(C) is the average number of instances in each category.}
    \label{table:statistic}
\end{table*}

Therefore, researching the few-shot ASQP task, i.e. \emph{fast adaptation to unseen aspects with only a few labeled samples}, becomes crucial, as it aligns more closely with real-world application scenarios. 
Yet, previous ASQP datasets either have a limited number of categories \cite{zhang2021aspect-sentiment} or long-tailed distribution \cite{cai2021aspect,zhang2021aspect-sentiment}. This task lacks a proper benchmark dataset. Therefore, we annotate a few-shot ASQP dataset, named $\mathtt{FSQP}$. This dataset aims to provide a more balanced representation and encompasses a wider range of categories, offering a comprehensive benchmark for evaluating few-shot ASQP. 


Recent studies have employed generative methods to extract quads by converting input sentences into templated target sequences \cite{zhang2021towards-generative,zhang2021aspect-sentiment,mao2022seq2path,bao2022aspect,peper2022generative,hu2022improving,hu2023uncertainty}. Subsequently, by disentangling the formats of template, quads can be extracted. However, they have primarily concentrated on the utilization of a single template \cite{zhang2021towards-generative} or incorporate multiple templates by considering different quad orders \cite{hu2022improving}, thereby ignore the correlation among these various templates. To overcome this limitation, we introduce an innovative method called Broad-view Soft Prompting (BvSP). BvSP leverages a pre-trained language model and utilizes Jensen-Shannon (JS) divergence to select several templates, enabling a 
more harmonious view of the available templates. We further introduce soft prompting to fine-tune the pre-trained language model with these selected templates. The final prediction is obtained from multiple templates by using a voting mechanism.

In summary, our major contributions of this paper are as follows:

\noindent
\begin{itemize}[leftmargin=*]
    \item We construct a new few-shot ASQP dataset $\mathtt{FSQP}$ which contains richer categories and is more balanced for the few-shot study. To the best of our knowledge, this is the first work to label the few-shot dataset in the ASQP task.
    \item We further propose BvSP, a various templates-based soft prompt learning method that improves quad prediction by taking into account the correlation between the different templates.
    \item Experimental results under four few-shot settings (i.e. one-shot, two-shot, five-shot, and ten-shot) demonstrate that BvSP outperforms strong baselines and has significant gains in other public datasets.
    
\end{itemize}

\begin{figure*}[t]
\centering
\subfigure[$\mathtt{Rest15}$]{
\includegraphics[width=0.15\textwidth]{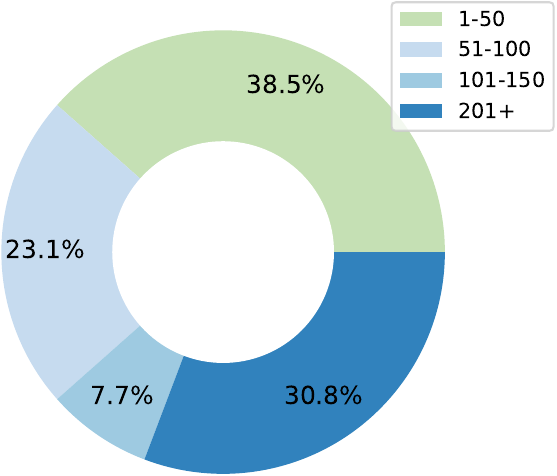}}
\hspace{0.5cm}
\subfigure[$\mathtt{Rest16}$]{
\includegraphics[width=0.15\textwidth]{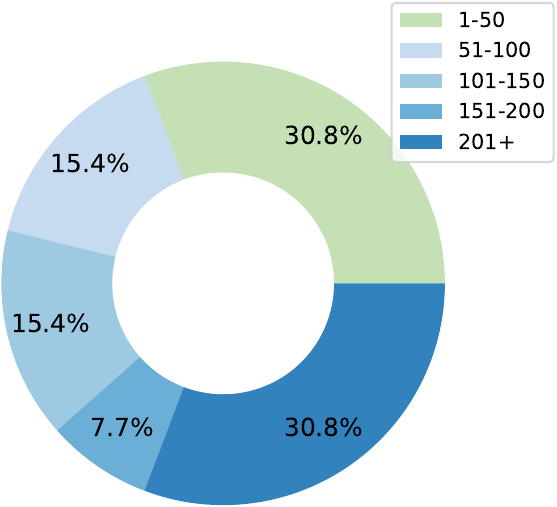}}
\hspace{0.5cm}
\subfigure[$\mathtt{Restaurant}$]{
\includegraphics[width=0.15\textwidth]{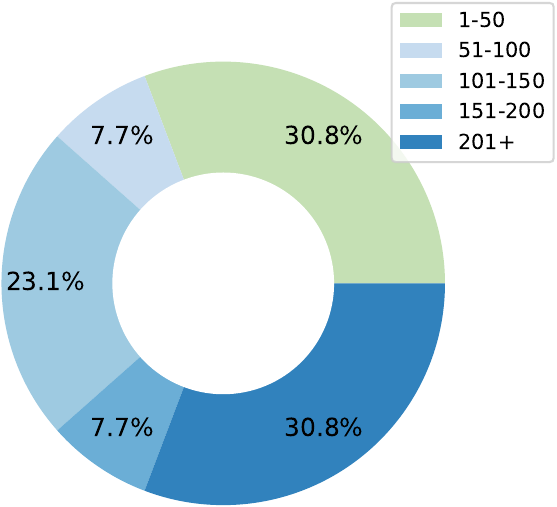}}
\hspace{0.5cm}
\subfigure[$\mathtt{Laptop}$]{
\includegraphics[width=0.15\textwidth]{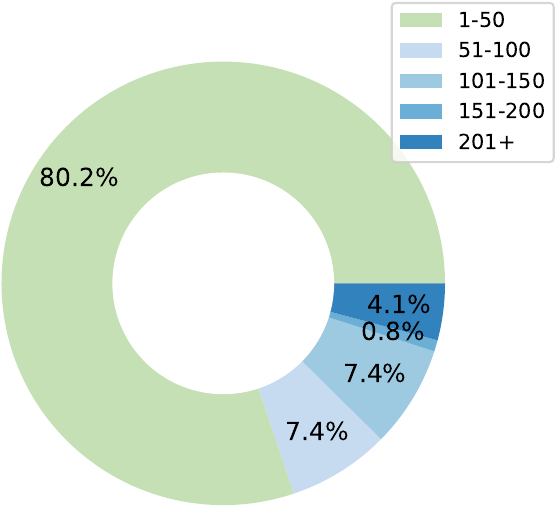}}
\hspace{0.5cm}
\subfigure[$\mathtt{FSQP}$]{
\includegraphics[width=0.15\textwidth]{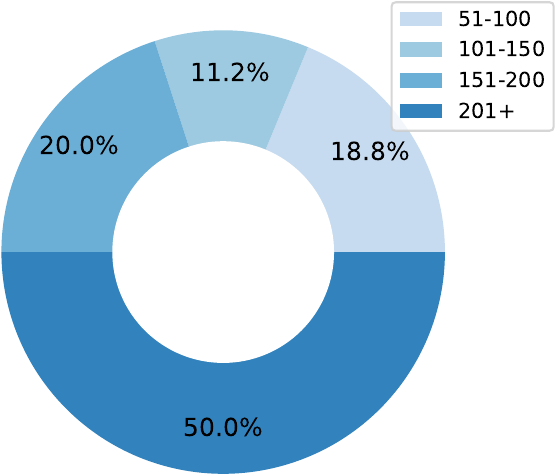}}
\caption{The category distribution is presented according to the number of instances. For example, the green section indicates the proportion of categories with the number of instances between 1 and 50.}
\label{fig:pie}
\end{figure*}

\section{Datasets}

To construct a dataset that is more representative of real-world scenarios, we manually annotate a new dataset, named \textbf{F}ew-\textbf{S}hot AS\textbf{QP} ($\mathtt{FSQP}$). 
In this section, we will first describe the data collection and annotation process. Following that, we will establish the superiority of $\mathtt{FSQP}$ by comparing it with previous datasets in terms of key statistics and features.
\subsection{Collection}

Our data source for this study is a collection of Yelp reviews spanning six years and originating from diverse cities across the United States. These reviews encompass various establishments, including restaurants, hotels, and beauty spas. Initially, these reviews were labeled with aspect categories and sentiment polarity by \citet{bauman2017aspect}. The $\mathtt{FSQP}$ dataset is an extension of this Yelp review data, featuring additional annotations and refinements.

\subsection{Annotation}

\subsubsection{Annotation Guidelines}

We consolidate the assessments across the three domains by aligning them with the same categories as per the evaluations carried out in prior works \cite{bauman2017aspect,li2019exploiting}. These categories encompass aspect categories and sentiment polarities, initially identified by the Opinion Parser (OP) system \cite{qiu2011opinion,liu2020sentiment}. The annotations generated by this system are manually double-checked for accuracy and completeness.
Afterward, we perform a thorough examination of the taxonomy and merge aspect categories that share similarities. Building on the double-checked aspect categories and sentiment polarity, we proceed to enrich the annotations of the \emph{aspect term} and the \emph{opinion term}.

Additionally, in line with the approaches proposed by \citet{cai2021aspect,poria2014rule}, we also consider the annotation of implicit \emph{aspect terms} and implicit \emph{opinion terms}.

\subsubsection{Annotation Process}

For selecting the aspect categories, the decision is made by a team of professional annotators. They check each category and its similarity one by one. For the annotation of other elements, 
two master's students who are well-versed in aspect-based sentiment analysis are chosen as annotators to annotate independently.
The strict quads matching F1 score between two annotators is 78.63\%, indicating a substantial agreement between them \cite{kim-klinger-2018-feels}. If one of the annotators disagrees with any content of the quad, they discuss to reach a consensus. Meanwhile, the leader of the master students will help to make the final decision. Throughout the six-month annotation period, while annotators were allowed to communicate with each other, their annotation operations remained relatively independent. Following this, there was a two-month period specifically set aside for verifying the accuracy of the annotations, during which we rigorously enforced consistency. Based on the above measures, the annotation of the new dataset has undergone careful scrutiny and discussion, suggesting a high level of credibility.

\begin{figure*}[t]
    \centering
    \includegraphics[width=0.80\textwidth]{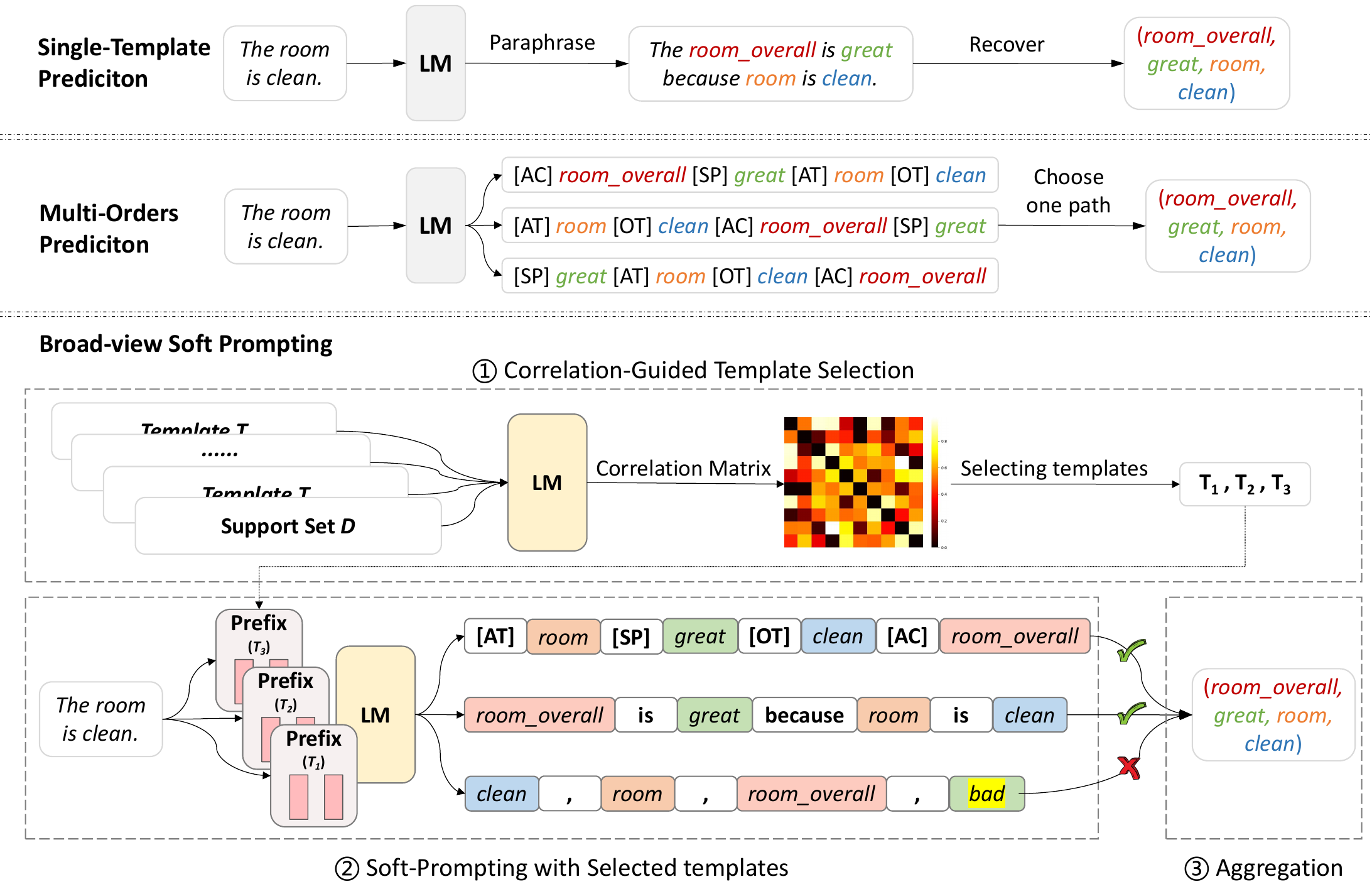}
    \caption{An overview of the proposed Broad-view Soft Prompting (BvSP). The single-template prediction is Paraphrase \cite{zhang2021towards-generative}. The multi-order prediction approach is DLO \cite{hu2022improving}. BvSP combines these templates as candidates and proposes a correlation-guided strategy for template selection.}
    \label{fig:overview}
\end{figure*}

\subsection{Statistics and Analyses}
\label{sec: data anal}
The statistics of $\mathtt{FSQP}$ can be found in Appendix \S\ref{sec:Supplementary Materials}. $\mathtt{FSQP}$ comprises 12,551 sentences, yielding a total of 16,383 quads. It's important to note that $\mathtt{FSQP}$ also encompasses implicit information, specifically, \emph{implicit aspect terms} and \emph{implicit opinion terms}, which are not explicitly mentioned in the sentence. Upon comparing these two types of implicit information, it becomes evident that the implicit opinion terms are more numerous than the implicit aspect terms.

In Table \ref{table:statistic}, we proceed with a more detailed comparison between $\mathtt{FSQP}$ and existing ASQP datasets regarding their distribution. It's evident from the table that $\mathtt{FSQP}$ surpasses even the current largest benchmark dataset, $\mathtt{Laptop}$, in terms of both scale and the number of quads. As a result, \textbf{$\mathtt{FSQP}$ stands out as a dataset with a more extensive collection of instances and quads.} Otherwise, based on the statistics of categories, it can be found that the existing benchmark dataset has insufficient categories and instances in each category. Our dataset effectively addresses both of these shortcomings. $\mathtt{FSQP}$ has a larger number of categories than the restaurant domain. Compared to $\mathtt{Laptop}$, it has a higher average number of instances in the category. 


The category distribution is also counted according to the number of instances in a certain interval. As shown in Figure \ref{fig:pie}, it can be seen that the categories with only 1-50 instances in previous benchmark datasets account for a majority percentage. Especially in the $\mathtt{Laptop}$, such categories account for 80.2\%. On the contrary, $\mathtt{FSQP}$ rarely has tailed categories. \textbf{Therefore, stable training and reliable testing can be achieved on $\mathtt{FSQP}$.} In summary, $\mathtt{FSQP}$ provides a suitable testbed to identify the capabilities of ASQP methods in real-world scenarios. It is worth noting that the design of $\mathtt{FSQP}$ follows the FewRel dataset \cite{han-etal-2018-fewrel}, which is proposed for few-shot relation classification. FewRel has a wide range of relation classes and balanced distribution, with 100 relations and each relation contains 700 instances.
\section{Methodology}

\subsection{Formulation and Overview}



Given a sentence $\bm{x}$ and its aspect sentiment quads $\{({at},{ot},{ac},{sp})\}$. Following the previous generation-based works \cite{zhang2021aspect-sentiment,hu2022improving}, we define projection functions to map the quads $({at},{ot},{ac},{sp})$ into semantic values $(x_{at},x_{ot},x_{ac},x_{sp})$. For example, we map the \emph{``POS''}, \emph{``NEU''}, and \emph{``NEG''} labels of sentiment polarity to \emph{``great''}, \emph{``ok''}, and \emph{``bad''}, and map the \emph{``NULL''} label of aspect term to \emph{``it''}. Based on the above rules, we use several templates, as shown in Figure \ref{fig:template}, to convert the aspect sentiment quadruples into target sequences that the language model can understand. If a sentence contains multiple quads, the target sequences are concatenated with a special marker $\mathtt{[SSEP]}$ to obtain the final target sequence.


As shown in Figure \ref{fig:overview}, we first use the pre-trained language model with JS divergence to select appropriate subsets of these templates, considering the efficiency and effectiveness of optimization, for a more detailed explanation, please refer to \S\ref{sec: more view}. Conditioned on soft prompts for different templates, it can generate multiple quads from diverse views. During the inference phase, different templates are generating results from different views. Thus, we aggregate these templates by voting for accurate quads extraction. Next, we will describe each section in detail.

\begin{figure}[t]
    \centering
    \includegraphics[width=0.48\textwidth]{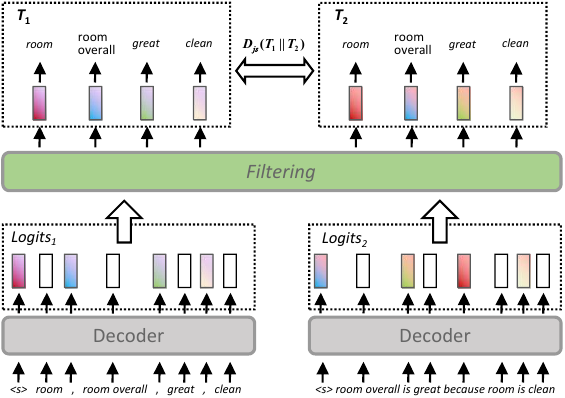}
    \caption{Quad elements (colored ones) of the output sequence from the decoder are filtered, which are leveraged for computing template correlations.}
    \label{fig:select}
\end{figure}

\subsection{Correlation-Guided Template Selection}

Though there are several available templates, jointly using them is inefficient. Thus, we choose templates by evaluating their correlations with a pre-trained language model (LM), where the detailed pre-training process is described in \S\ref{sec:detail}. In this way, chosen templates are more suitable for the nature of LM. Concretely, the correlation between the two templates is based on the whole support set $\mathcal{D}$. We compute the average score across all instances of these two templates. Specifically, given an input $\bm{x}$, its quads, and several available templates $\mathcal{T}$ = \{$\mathcal{T}_1$, $\mathcal{T}_2$, ..., $\mathcal{T}_{T}$\} where $T$ is the number of templates, we use $\bm{y^{i}}$ and $\bm{y^{j}}$ to indicate the target sequences under any two templates $\mathcal{T}_i$ and $\mathcal{T}_j$. The target sequence is usually composed of quad elements and linking symbols. As shown in Figure \ref{fig:select}, for both two templates, (\emph{``room''}, \emph{``room\_overall''}, \emph{``great''}, \emph{``clean''}) denote the quad elements. The \emph{``,''}, \emph{``is''}, and \emph{``because''} indicate symbols linking the quad.

Therefore, we can choose the quad part from each target sequence. We denote this process by the following equation.
\begin{equation}
\bm{h^{i}}=\mathrm{Filtering}(\bm{x}, \bm{y^{i}})
    \label{eq:filter-reorder}
\end{equation}
where $\bm{h^{i}}$ is set of selected representations from the target sequence and $\bm{y}^{i}$ is fed into the decoder as teacher forcing \cite{williams1989learning}. Then, we obtain the correlation of two target sequences by using JS divergence.
\begin{align}
\small
    \begin{split}
    \notag
    &\mathcal{D}_{KL}(\bm{p}||\bm{q})=\sum_{|V|}{\bm{p}}\mathrm{log}(\frac{\bm{p}}{\bm{q}})
    \end{split}
    \\
\small
    \begin{split}
    \mathcal{D}_{JS}(\bm{h^{i}}||\bm{h^{j}})=\frac{1}{|\bm{h^{i}}|}\sum_{|\bm{h^{i}|}}(\frac{1}{2}\mathcal{D}_{KL}(\bm{h^{i}}||\frac{\bm{h^{i}}+\bm{h^{j}}}{2}) \\
    +\frac{1}{2}\mathcal{D}_{KL}(\bm{h^{j}}||\frac{\bm{h^{i}}+\bm{h^{j}}}{2}))
    \end{split}
\end{align}
where $|V|$ is the size of the vocabulary set and $|\bm{h^{i}}|$ is the number of selected quad tokens. The $\mathcal{D}_{KL}$ represents the calculation of the KL divergence of two probability distributions.

For the whole support set $\mathcal{D}$, we have $(\bm{x},\{\bm{y}^{i}\}_{i=1}^{T})$ for each instance by constructing templates. Then the average score of templates is calculated over the support set:
\begin{equation}
    \mathcal{S}_{\mathcal{T}_i, \mathcal{T}_j}=\frac{1}{|\mathcal{D}|}\sum_{\mathcal{D}}\mathcal{D}_{JS}(\bm{h^{i}}||\bm{h^{j}})
\end{equation}
where $\mathcal{S}_{\mathcal{T}_i, \mathcal{T}_j}$ denotes the average correlation of all instances between templates $\mathcal{T}_i$ and $\mathcal{T}_j$.

Then with the correlation between every two templates, we can obtain a correlation matrix $\mathcal{S} \in \mathbb{R}^{|T|\times|T|}$. Then, we enumerate the entire matrix and find the $k$ matrix points with the smallest values. These points are the most relevant $k$ templates while are our final fine-tuned templates. We aggregate multiple templates by taking into account their shared quads. When the templates exhibit greater disparity, conflicts arise, leading to inconsistent quads and empty predictions. Conversely, when the correlation between templates is stronger, they are more harmonious, fostering consistent support for the quad.

\subsection{Soft-Prompting with Selected Templates}

Then we aim to incorporate various selected templates. Yet it is difficult to distinguish between them. Thus we design a specific prefix for each template that can act as an indicator for each of them. With the input $\bm{x}$, the target sequence $\bm{y}_t$, and the succession of prefix parameters $\bm{z^{i}}$ for the template $\mathcal{T}_i$ \cite{li2021prefix}, we fine-tune with minimizes cross-entropy loss defined as follows:
\begin{equation}
    \mathcal{L}(\bm{x},\bm{y^{i}})=-\sum_{t=1}^n\mathrm{log}p_\theta(\bm{y}^i_t|\bm{x},\bm{z^{i}},\bm{y}^i_{<t})
\end{equation}
where $n$ is the length of the target sequence $\bm{y^{i}}$.

\tabcolsep=0.5cm
\begin{table*}[!t]
\renewcommand\arraystretch{1.4}
\small
    \centering
    
    \setlength{\tabcolsep}{1.2mm}{\begin{tabular}{l ccc ccc ccc ccc ccc}
    \toprule
    \multirow{2}{*}{Baseline} & \multicolumn{3}{c}{One-shot} & \multicolumn{3}{c}{Two-shot} & \multicolumn{3}{c}{Five-shot} & \multicolumn{3}{c}{Ten-shot} & \multicolumn{3}{c}{Avg} \\
    & $\mathtt{Pre}$ & $\mathtt{Rec}$ & $\mathtt{F1}$ & $\mathtt{Pre}$ & $\mathtt{Rec}$ & $\mathtt{F1}$ & $\mathtt{Pre}$ & $\mathtt{Rec}$ & $\mathtt{F1}$ & $\mathtt{Pre}$ & $\mathtt{Rec}$ & $\mathtt{F1}$ & $\mathtt{Pre}$ & $\mathtt{Rec}$ & $\mathtt{F1}$ \\
    \midrule
    GAS & \underline{26.39} & \underline{26.70} & \underline{26.54} & 40.42 & 41.21 & 40.81 & 49.70 & 49.89 & 50.50 & 51.89 & 52.80 & 52.34 & 42.10 & 42.65 & 42.55 \\
    Paraphrase & 23.53 & 23.25 & 23.38 & 39.04 & 38.92 & 38.98 & 49.04 & 49.54 & 49.29 & 52.02 & 52.52 & 52.26 & 40.91 & 41.06 & 40.97 \\
    DLO & 26.37 & 26.14 & 26.23 & 36.70 & 38.80 & 37.72 & 49.05 & \textbf{\underline{51.80}} & 50.38 & 50.48 & 52.83 & 51.62 & 40.65 & 42.39 & 41.48 \\
    ILO & 25.47 & 24.35 & 24.98 & 37.71 & 38.54 & 38.12 & \underline{50.64} & 51.53 & \underline{50.95} & 50.52 & 53.04 & 51.94 & 41.08 & 41.87 & 41.50 \\
    MvP & 25.82 & 25.86 & 25.84 & \underline{40.64} & \underline{41.65} & \underline{41.14} & 49.05 & 50.43 & 49.73 &  \underline{53.05} & \textbf{\underline{54.18}} & \underline{53.61} & \underline{42.14} & \underline{43.03} & \underline{42.58} \\ 
    
    \midrule
    ChatGPT & 23.79 & 23.24 & 23.56 & 33.24 & 38.14 & 35.52 & 36.29 & 40.63 & 38.97 & 46.03 & 40.25 & 42.96 & 34.84 & 35.56 & 35.25 \\
    \midrule
    BvSP & \textbf{45.50} & \textbf{32.62} & \textbf{37.99} & \textbf{52.12} & \textbf{43.41}  & \textbf{47.34} & \textbf{56.87} & 50.55 & \textbf{53.52} & \textbf{57.29} & 52.04 & \textbf{54.53} & \textbf{52.95} & \textbf{44.66} & \textbf{48.35} \\
    \bottomrule
    \end{tabular}}
    \caption{Evaluation results of few-shot ASQP task, compared with baseline methods in terms of precision ($\mathtt{Pre}$, \%), recall ($\mathtt{Rec}$, \%) and F1 score ($\mathtt{F1}$, \%). 
    Under each column, the best results are marked in bold and the best baseline results are underlined.}
    \label{table:asqp_results_}
\end{table*}

\subsection{Multi-Templates Aggregation}

During the inference phase, following \cite{gou2023mvp}, BvSP aggregates the quad results generated by all selected templates. Subsequently, we employ a voting mechanism to determine the quad that garners the high frequency of predictions and designate it as the final prediction. By defining a threshold $\tau$, the quad is extended to the final prediction when the number of votes for this quad is larger than this threshold.
\begin{equation}
\notag
\mathcal{P}=\{q|q \in \bigcup_{i=1}^k\mathcal{T}_{i} \text{ and } (\sum_{i=1}^k \text{$\mathbbm{1}$}_{\mathcal{T}_{i}}(q)\geq \tau)\}
\end{equation}
where $q$ denotes the quad obtained in the template $\mathcal{T}_{i}$ and $\mathcal{P}$ is the final prediction of quads.

\section{Experiments}

\subsection{Implementation Details}
\label{sec:detail}
 In the experiments, all the reported results are the average of 5 runs. We adopt T5-base \cite{JMLR:v21:20-074} as the pre-trained generative model. For all baselines and our method, they are pre-trained first on the training set and then fine-tuned with the support set. Our work follows the common few-shot setting, where regardless of the number of shots, all samples from the training set are used to train the model. The \emph{``k-shot''} concept refers to the number of labeled instances provided for each class during the few-shot task. For example, in a 1-shot setting, only one labeled instance is available per class in the support set, while in a 2-shot setting, two labeled instances are available per class. In our experiments, for k-shot, we randomly sample k instances in the test set as the support set and the remaining instances in the test set as the query set. Especially, BvSP uses all templates to pre-train the prefix parameters on the training set. It is worth noting that in BvSP we freeze the LLM parameters and only fine-tune the parameters of the prefix. 

In addition, we also depict the template details of each baseline, more details about the templates are given in Appendix \S\ref{sec:Template details of various methods}. During the pre-training phase, we set the epoch to 20, batch size to 16, and learning rate to 3e-4, except for DLO, ILO, and MvP, where the learning rate is set to 1e-4. For the support sets in the fine-tuning stage, we maintain the epoch at 20, choose a batch size of 8, and utilize a learning rate of 3e-4. During the inference stage, all methods employ a beam size of 1, except for DLO and ILO, which use a beam size of 5. 

\subsection{Datasets}

The original dataset $\mathtt{FSQP}$ is first divided into a training set, a development set, and a testing set according to aspect category.
More details about datasets are given in Appendix \S\ref{sec:Datasets statistics}.
Few-shot learning aims to replicate real-world scenarios where the model encounters novel classes not included in the training dataset. Consequently, these three datasets have varying numbers of aspect categories with no overlap.
We further conduct the experiments on our dataset $\mathtt{FSQP}$ under the four few-shot settings.

\subsection{Compared Methods}

We choose the strong generative baseline methods. They include both the currently most popular LLMs, i.e. ChatGPT \cite{brown2020language}, as well as the state-of-the-art (SOTA) methods in sentiment analysis in recent years, namely GAS \cite{zhang2021towards-generative}, Paraphrase \cite{zhang2021aspect-sentiment}, DLO, ILO \cite{hu2022improving}, and MvP \cite{gou2023mvp}.

    
    
    

\subsection{Experimental Results}

Experimental results of the ASQP task under few-shot settings are reported in Table \ref{table:asqp_results_}. 
For a fair comparison, we follow the settings of \citet{hu2022improving} and set the default number of selected templates to top-3. Our method BvSP still has room for improvement in the performance (see \S\ref{sec: more view}).


It can be seen that BvSP achieves the best performance in the four few-shot settings. Especially, compared with baseline GAS, BvSP gains absolute F1 score improvement by +11.45\% under the one-shot setting. Moreover, BvSP is able to outperform DLO and ILO with the same number of selected templates under the same few-shot settings. Under the same few-shot settings, BvSP demonstrates superior performance compared to MvP. These results validate the effectiveness of BvSP in providing a broader view of templates.

\begin{table}[]
\renewcommand\arraystretch{1.3}
\small
    \centering

    \setlength{\tabcolsep}{2.5mm}{\begin{tabular}{ll ccc}
    \toprule
    \multicolumn{2}{c}{\multirow{2}{*}{Method}} & \multicolumn{3}{c}{Ten-shot} \\
    && $\mathtt{Pre}$ & $\mathtt{Rec}$ & $\mathtt{F1}$ \\
    \midrule
    & BvSP (JS Min) & \textbf{57.29} & \textbf{52.04} & \textbf{54.53} \\
    \midrule
    \multirow{4}{*}{\textbf{I}} & BvSP (JS Max) & 56.29 & 50.09 & 53.01 \\
    & BvSP (Entropy Min) & 62.90 & 44.30 & 51.98 \\
    & BvSP (Entropy Max) & 61.42 & 45.67 & 52.19 \\
    & BvSP (random) & 56.27 & 50.27 & 53.10 \\
    \hdashline
    \multirow{2}{*}{\textbf{II}}&BvSP (rank) & 56.29 & 50.09 & 53.01 \\
    & BvSP (rand) & 54.27 & 51.38 & 52.78 \\
    \bottomrule
    \end{tabular}}
    \caption{Evaluation results of ablation study.}
    \label{table:ablation study}

\end{table}

\subsection{Ablation Study}

To demonstrate the effectiveness of BvSP in selecting templates and aggregation methods, ablation experiments are performed. The results are shown in Table \ref{table:ablation study}. Following the default setting of BvSP, the model variants also select top-3 templates. The model variants first study from (\textbf{I}) template selection strategy, including maximum and minimum entropy, maximum JS divergence, i.e. BvSP (JS Max), random sampling, i.e. BvSP (random). During inference, we further study the (\textbf{II}) aggregation strategies, including BvSP (rank) and BvSP (rand). The former selects the top-ranked sequence by considering the perplexity of the generated sequences, while the latter selects one sequence randomly.

It is first observed that using the minimal JS divergence consistently outperforms the other strategies. Specifically, compared with BvSP (JS Max), BvSP (Entropy Min), BvSP (Entropy Max), and BvSP (random), BvSP makes absolute F1 score improvements by +1.52\%, +2.55\%, 2.34\%, and +1.43\%, respectively. These results underscore the effectiveness of our strategy. Selecting correlated templates can make them cooperate harmoniously.



Furthermore, BvSP outperforms BvSP (rank) and BvSP (rand) by +1.52\% and +1.75\% F1 score. Our aggregation method using voting demonstrates superiority over the remaining two strategies. Actually, both of these strategies produce outputs from a single template, which could be either random or ranking selection. In contrast to these methods, we select the final prediction by considering the commonalities between multiple templates. 

\begin{figure}[t]
\centering
\includegraphics[width=0.47\textwidth]{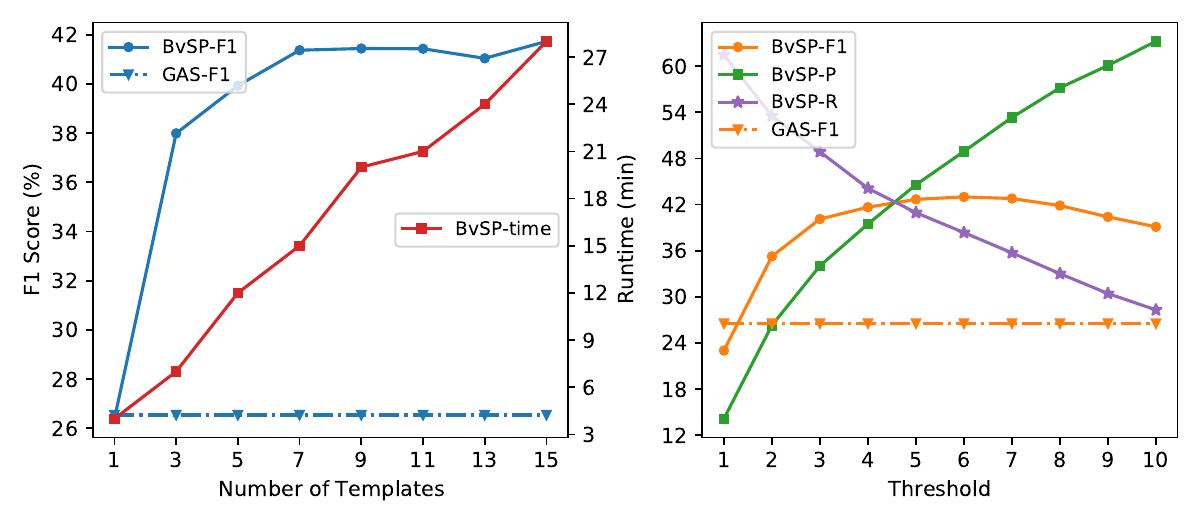}

\caption{Effects of hyperparameters on $\mathtt{FSQP}$ under the one-shot settings.}
\label{pic:hyperparameters}
\end{figure}

\subsection{Hyperparameter Study}
\label{sec: more view}

The study includes an examination of the effects of two hyperparameters: $k$ and $\tau$. The $k$ represents the number of selected templates, while $\tau$ refers to the threshold for gaining final quads from multiple templates. Figure \ref{pic:hyperparameters} illustrates the resulting curves.

\noindent
\textbf{Number of selected templates $\boldsymbol{k}$}: This hyperparameter affects the breadth of templates view. Here we set various $k$ values and also change the threshold $\tau=k/2$ accordingly. All other hyperparameters are kept the same. Analyzing the left plot of Figure \ref{pic:hyperparameters}, it is seen that BvSP outperforms the baseline GAS in most cases, indicating the robustness of this hyperparameter to some extent. 
While a large $k$ increases time consumption, opting for a smaller value results in reduced performance. The figure clearly demonstrates that when considering both the performance of BvSP and the time consumed, the optimal choice is $k=3$.

\noindent
\textbf{Threshold $\boldsymbol{\tau}$}: This hyperparameter determines the confidence level of the quads in the final prediction $\mathcal{P}$. A higher threshold signifies that the quads predicted by the model are more frequently matched by multiple templates, thereby enhancing confidence in the correctness of the prediction of final quads. 
In the right plot of Figure \ref{pic:hyperparameters}, we observe that BvSP keeps increasing in accuracy and decreasing in recall as $\tau$ increases and produces optimal F1 score by maintaining within the range of 5 to 7. This indicates that setting $\tau$ to a larger value results in fewer screened quads but more evidence for each selected quad. 

\tabcolsep=0.5cm
\begin{table}[!t]
\renewcommand\arraystretch{1.3}
\small
    \centering
    
    \setlength{\tabcolsep}{1.2mm}{\begin{tabular}{l ccc ccc}
    \toprule
    \multirow{2}{*}{Baseline} & \multicolumn{3}{c}{$\mathtt{Rest15}$} & \multicolumn{3}{c}{$\mathtt{Rest16}$} \\
    & $\mathtt{Pre}$ & $\mathtt{Rec}$ & $\mathtt{F1}$ & $\mathtt{Pre}$ & $\mathtt{Rec}$ & $\mathtt{F1}$  \\
    \midrule
    GAS & 45.31 & 46.70 & 45.98 & 54.54 & 57.62 & 56.04 \\
    Paraphrase & 46.16 &	47.72 &	46.93 &	56.63 &	59.30 &	57.93 \\
    DLO & 47.08 &	49.33 &	48.18 &	57.92 &	\textbf{61.80} &	59.79 \\
    ILO & 47.78 & \textbf{50.38} & 49.05 &	57.58 &	61.17 &	59.32 \\
    MvP (top-3) & 49.59 &	48.93 &	49.32 &	60.27 &	58.94 &	59.62 \\
    MvP (top-15) & - & - & 51.04 & - & - & 60.39 \\
    \midrule
    BvSP (top-3) & 54.63 & 47.53 & 50.83 & 63.59 & 59.35 & 61.40 \\
    BvSP (top-15) & \textbf{60.96} & 47.15 & \textbf{53.17} & \textbf{68.16} & 59.42 & \textbf{63.49} \\
    \bottomrule
    \end{tabular}}
    \caption{Full-shot results of ASQP task in the datasets $\mathtt{Rest15}$ and $\mathtt{Rest16}$.}
    \label{table:asqp_results}
\end{table}

\subsection{Evaluation on Other Datasets}

We perform the ASQP task on the Rest15 and Rest16 datasets using baselines and the proposed BvSP in the full-shot setting. The results are presented in Table \ref{table:asqp_results}. BvSP continues to outperform all baselines in the full-shot setting. It is worth noting that BvSP exhibits superiority over MvP in both the top-3 and top-15. This underscores that BvSP not only excels in few-shot scenarios but also proves beneficial in full-shot scenarios. 

\subsection{Case Study}

\begin{figure}[t]
\centering
\includegraphics[width=0.48\textwidth]{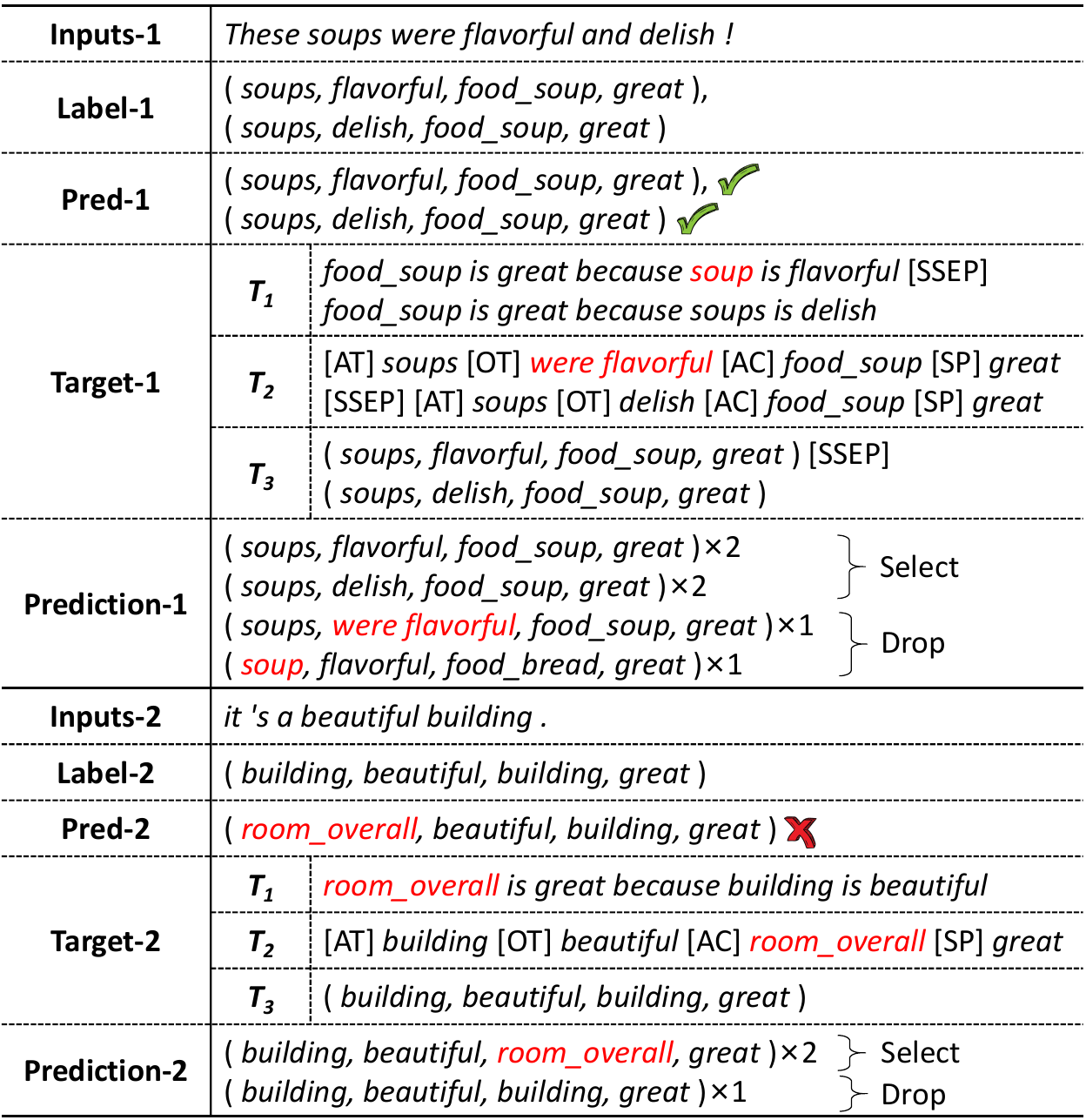} 
\caption{Two cases predicted by BvSP from the testing set of $\mathtt{FSQP}$ dataset under the one-shot settings.}  
\label{fig:error_anal}
\end{figure}

To completely understand the strengths and weaknesses of the BvSP, we conduct a case study. We provide one instance that demonstrates a successful prediction, while another instance showcases an error prediction by our method. These two cases are presented in Figure \ref{fig:error_anal}.


In case 1, we can find that two of the three templates have prediction errors (marked in red color). However, after passing our voting mechanism, the final prediction is corrected. This indicates clearly that BvSP shows a remarkable ability to successfully aggregate exactly predicted quads from multiple templates while simultaneously filtering out the error predictions among them by considering the commonalities between templates. Besides, in case 2, two out of the three templates are erroneously predicted (marked in red color), subsequently leading to an error final prediction. This case signifies that the accuracy of our methodology remains dependent on the rationality of a template. If multiple templates contain the same kind of error, this is hardly eliminated error by the voting mechanism. Thus, the observation highlights the fact that our aggregation strategy still possesses untapped potential for enhancement. 


\section{Related Work}

\subsection{Datasets for ASQP}

The current ASQP datasets \cite{cai2021aspect,zhang2021aspect-sentiment} are generated through similar processes and have a shared origin. Specifically, they are derived from SemEval challenge datasets \cite{pontiki-etal-2014-semeval,pontiki2015semeval,pontiki2016semeval}, which provide aspect terms and corresponding sentiments for restaurant and laptop reviews. \citet{fan2019target} annotate the aspect-opinion pairs based on the SemEval datasets, and \citet{peng2020knowing} further expand sentiment tuples $(at, ot, sp)$ based on \cite{fan2019target}. \citet{zhang2021aspect-sentiment} annotate $\mathtt{Rest15}$ and $\mathtt{Rest16}$ in light of the Semval tasks \cite{pontiki2015semeval,pontiki2016semeval}. Meanwhile, \citet{cai2021aspect} propose $\mathtt{Restaurant}$ and $\mathtt{Laptop}$. $\mathtt{Restaurant}$ is based on initial SemEval challenge datasets and its extensions \cite{fan2019target,xu2020position}, while \citet{cai2021aspect} annotate the $\mathtt{Laptop}$ dataset based on the Amazon 2017 and 2018 reviews.

However, all of these datasets still contain many limitations, as explained in \S\ref{sec: data anal}. 
The shortcomings of existing benchmark datasets motivate us to provide a more diverse set of reviews covering more categories and instances of each category. 

\subsection{Methods for ASQP}

Aspect Sentiment Quad Predicition (ASQP) has received wide attention in recent years, learning four elements simultaneously, i.e. aspect sentiment quads. Corresponding solutions for ASQP can be divided into two categories: non-generation \cite{cai2021aspect} and generation \cite{zhang2021aspect-sentiment,mao2022seq2path,bao2022aspect,peper2022generative,hu2022improving,hu2023uncertainty,gou2023mvp}. The non-generation method employs two-stage framework to extract aspect sentiment quads by improving tradition-based methods such as Double-Propagation \cite{10.1162/coli_a_00034}, JET \cite{xu2020position}, HGCN \cite{zhang2021aspect-sentiment}, TAS \cite{wan2020target}. Due to the simplicity and end-to-end manner, generation-based methods have become the main research direction. Promising works design novel approaches based on tree structure \cite{mao2022seq2path,bao2022aspect}, contrastive learning \cite{peper2022generative}, data augmentation \cite{hu2022improving}, multi-view \cite{gou2023mvp} and uncertainty \cite{hu2023uncertainty}. In contrast to these works, we approach the few-shot ASQP task from a broader perspective, introducing Broad-View Soft Prompting (BvSP) into our design.

\section{Conclusion}


This work studies the ASQP task from the few-shot perspective, which aims to handle unseen aspects with only a few supported samples. Therefore, we first build a new dataset called $\mathtt{FSQP}$, which is specifically annotated for few-shot ASQP. Unlike existing ASQP datasets, $\mathtt{FSQP}$ provides a more balanced representation and covers a wider range of categories, thereby serving as a comprehensive benchmark for evaluating few-shot ASQP. Moreover, the generation-based paradigm has become the state-of-the-art technique for ASQP. However, previous methods overlook the correlation between different templates. In this study, we propose a broad-view soft prompting (BvSP) method to address this limitation. BvSP leverages the JS divergence to analyze the correlation among templates and selects relevant ones. It then guides the pre-trained language model with soft prompts based on these selected templates. Finally, the results are aggregated through voting. Extensive experiments conducted under few-shot settings demonstrate that BvSP exhibits universal effectiveness and substantial improvements in both explicit and implicit information.


\section*{Limitations}

Our work represents the pioneering effort in tackling the few-shot ASQP task through the creation of a novel benchmark dataset, $\mathtt{FSQP}$, and the introduction of a novel method, BvSP. However, it's essential to acknowledge that our work still has limitations, which can serve as valuable pointers for future research directions.

Firstly, we employ JS divergence to analyze the correlation between different templates. A smaller JS value signifies a stronger correlation between the two templates. Nonetheless, there may exist alternative criteria for template selection that could further boost the pre-trained language model, enhancing its support for the few-shot ASQP task. 

Secondly, the experiments focus solely on the few-shot ASQP. However, compound ABSA includes numerous subtasks, such as aspect category sentiment analysis \cite{schmitt-etal-2018-joint} and target aspect sentiment detection \cite{wan2020target}. 
Nevertheless, the $\mathtt{FSQP}$ dataset still satisfies the requirements for these tasks.
Therefore, future research may consider exploring few-shot learning techniques for these tasks.

Lastly, it should be noted that utilizing multi-soft prompting introduces additional training and inference overheads, which scale proportionally with the number of selected templates. Despite this, our method relies solely on automatic optimization and does not raise human labor.

\section*{Acknowledgements}
We sincerely thank all the anonymous reviewers for providing valuable feedback. This work is supported by the youth program of National Science Fund of Tianjin, China (Grant No. 22JCQNJC01340).

\bibliographystyle{acl_natbib}
\bibliography{my_refer}

\appendix
\clearpage

\section{Datasets statistics}\label{sec:Datasets statistics}

The original dataset $\mathtt{FSQP}$ is first divided into a training set, a development set, and a testing set according to aspect category.
The statistics are displayed in Table \ref{table:data}. 
Few-shot learning aims to replicate real-world scenarios where the model encounters novel classes that were not included in the training dataset. Therefore, it can be observed that three sets have 45, 15, and 20 aspect categories without overlapping. 
\begin{table}[]
\small
    \centering
    
    \begin{tabular}{l c c c}
    \toprule
     Datasets & \#S & \#Q & \#C \\
    \midrule
    Train & 8204 & 11363 & 45 \\
    Dev & 2204 & 2600 & 15 \\
    Test & 2143 & 2420 & 20 \\
\bottomrule
    \end{tabular}
    \caption{Data statistics. \#S, \#Q, and \#C denote the number of sentences, quads, and aspect categories respectively.}
    \label{table:data}
\end{table}

\section{Template details of various methods}\label{sec:Template details of various methods}
\begin{figure}[t]
\centering
\includegraphics[width=0.48\textwidth]{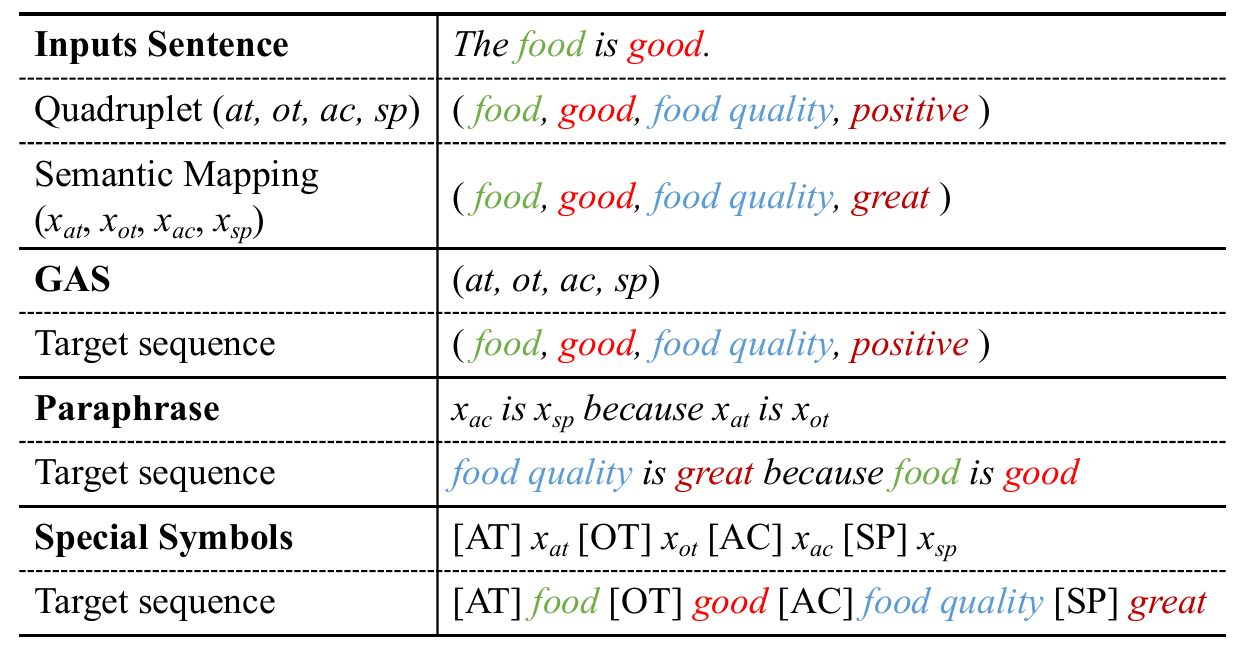} 
\caption{Template details of various methods.}
\label{fig:template}
\end{figure}

The template details of each baseline method in Figure \ref{fig:template}.
It is worth noting that ILO and DLO also follow the special symbols templates but combine multiple template orders as data augmentation. For BvSP, we consider all possible templates, including one for GAS \cite{zhang2021towards-generative}, one for Paraphrase \cite{
zhang2021aspect-sentiment}, and twenty-four for Special Symbols templates \cite{hu2022improving} including all possible permutations, thus twenty-six in total.

\begin{table}[]
    \centering
    \newcommand{\tabincell}[2]{\begin{tabular}{@{}#1@{}}#2\end{tabular}}
    \scalebox{0.60}{\setlength{\tabcolsep}{2.0mm}{\begin{tabular}{l l c c c c c}
    \toprule
     Element & Method & One-shot & Two-shot & Five-shot & Ten-shot & Average \\
    \midrule
    \multirow{5}{*}{\tabincell{l}{Aspect\\Term}} & GAS & 79.44 & 81.64 & 84.04 & 85.42 & 82.63 \\
    & Para & 79.64 & 82.73 & 85.41 & 86.77 & 83.63 \\
    & DLO & 71.33 & 80.32 & 85.32 & 87.40 & 81.09 \\
    & ILO & \textbf{80.18} & 82.45 & 84.36 & 86.64 & 83.40 \\
    \hdashline
    & BvSP & 79.41 & \textbf{82.66} & \textbf{85.40} & \textbf{88.15} & \textbf{83.90} \\
    \midrule
    \multirow{5}{*}{\tabincell{l}{Opinion\\Term}} & GAS & 72.73 & 73.50 & 74.55 & 74.97 & 73.93 \\
    & Para & 72.41 & 73.69 & 74.47 & 74.88 & 73.86 \\
    & DLO & 64.99 & 70.39 & 73.96 & 74.66 & 71.00 \\
    & ILO & 73.35 & \textbf{74.11} & 74.80 & \textbf{75.20} & \textbf{74.36} \\
    \hdashline
    & BvSP & \textbf{73.63} & 73.62 & \textbf{74.85} & 74.75 & 74.21 \\
    \midrule
    \multirow{5}{*}{\tabincell{l}{Aspect\\Category}} & GAS & 43.99 & 69.17 & 83.75 & 86.60 & 70.87 \\
    & Para & 38.88 & 65.65 & 82.85 & 86.37 & 68.43 \\
    & DLO & 46.11 & 64.94 & 85.34 & 87.60 & 70.99 \\
    & ILO & 38.90 & 60.99 & 83.05 & 86.98 & 67.48 \\
    \hdashline
    & BvSP & \textbf{62.00} & \textbf{80.08} & \textbf{86.97} & \textbf{87.93} & \textbf{79.25} \\
    \midrule
    \multirow{5}{*}{\tabincell{l}{Sentiment\\Polarity}} & GAS & 80.63 & 81.71 & 81.76 & 82.27 & 81.59 \\
    & Para & 80.63 & 81.21 & 81.18 & 82.06 & 81.27 \\
    & DLO & 73.06 & 78.63 & 81.21 & 81.99 & 78.72 \\
    & ILO & \textbf{81.69} & 81.70 & 81.76 & 82.33 & 81.87 \\
    \hdashline
    & BvSP & 80.71 & \textbf{82.05} & \textbf{82.74} & \textbf{84.48} & \textbf{82.50} \\
    \bottomrule
    \end{tabular}}}
    \caption{Analyses at aspect-based elements for few-shot ASQP in terms of F1 score (\%).}
    \label{table:analysis_aspect_based_asqp}
\end{table}

\section{Supplementary Experiments}

\subsection{Analysis at Element-Level}
In this section, the performance of BvSP is fully analyzed at four aspect-level elements: aspect term, opinion term, aspect category, and sentiment polarity. The corresponding analysis results are presented in Table \ref{table:analysis_aspect_based_asqp}.

Considering the aspect term and sentiment polarity first, BvSP shows superior performance compared to the baselines on the average F1 score. In addition, BvSP consistently surpasses the baselines in four few-shot scenarios under the aspect category. However, for the opinion term, our method only slightly less-performs the strong baseline ILO on the average F1 score. These indicate the effectiveness and robustness of BvSP on four elements from the unseen category.

Moreover, it is noticed that all methods achieve better performance in predicting aspect term, opinion term, and sentiment polarity in the one-shot scenario (more than 70\% F1 scores), but they have poor performance for the aspect category (only around 40\% F1 scores). This finding indicates that even in unseen aspect categories scenarios, the model can still effectively reason about the aspect term, the opinion term, and the sentiment polarity based on the knowledge and patterns acquired during the pre-training phase. A possible reason is that sentences share similar grammar rules and syntactic structures. Yet comprehending the semantics of unseen categories is struggling. 

Finally, comparing from 1 to 10 shots, the performance improvement in the aspect category is more significant than the other three elements. This possibly shows that feeding neural networks the novel knowledge contributes more to their fast adaptation, pointing out the potential future direction.

\subsection{Vote Mechanism}
To investigate the effect of the aggregation method, we conducted further comparative experiments on the voting mechanism, comparing BvSP with the optimal baseline, DLO, which also utilizes multiple templates. The results of these comparisons are presented in Table \ref{table:vote mechanism}. 
We can see that DLO (vote) performs better than DLO (one path), suggesting that employing the voting mechanism instead of choosing one path can enhance DLO's performance. However, while DLO exhibits some performance improvement, it still doesn't surpass our proposed BvSP method, further underscoring the effectiveness of our approach.

\begin{table}[]
\renewcommand\arraystretch{1.3}
\small
    \centering
    \setlength{\tabcolsep}{1mm}{\begin{tabular}{l c c c c c}
    \toprule
      & Method & One-shot & Two-shot & Five-shot & Ten-shot \\
    \midrule
    &BvSP & \textbf{37.99} & \textbf{47.34} & \textbf{53.52} & \textbf{54.43} \\
    \midrule
    &DLO (one path) & 26.23 & 37.72 & 50.38 & 51.62\\
    &DLO (vote) & 27.91 & 38.60 & 50.81 & 51.79 \\
    \bottomrule
    \end{tabular}}
    \caption{Evaluation results ($\mathtt{F1}$, \%) of vote mechanism.}
    \label{table:vote mechanism}

\end{table}

\begin{table}[]
\small
    \centering
    \setlength{\tabcolsep}{1mm}{\begin{tabular}{l c c c c c}
    \toprule
      & Method & One-shot & Two-shot & Five-shot & Ten-shot \\
    \midrule
    \multirow{5}{*}{Explicit} & GAS & 26.74 & 41.13 & 50.55 & 52.67 \\
    & Para & 24.16 & 39.40 & 51.19 & 53.69 \\
    & DLO & 26.33 & 38.57 & 51.44 & 52.82 \\
    & ILO & 25.41 & 37.53 & 50.74 & 53.09 \\
    \hdashline
    & BvSP & \textbf{41.66} & \textbf{49.73} & \textbf{56.22} & \textbf{56.77} \\
    \midrule
    \multirow{5}{*}{Implicit} & GAS & 23.01 & 37.13 & 46.32 & 48.66 \\
    & Para & 21.45 & 37.59 & 44.71 & 48.63 \\
    & DLO & 25.72 & 35.43 & 47.44 & 48.52 \\
    & ILO & 20.32 & 33.16 & 46.38 & 48.93 \\
    \hdashline
    & BvSP & \textbf{31.53} & \textbf{49.05} & \textbf{48.45} & \textbf{51.44} \\
    \bottomrule
    \end{tabular}}
    \caption{Evaluation results on the explicit and implicit subsets in terms of F1 score (\%).}
    \label{table:implicit information}
\end{table}

\subsection{Implicit Information Prediction}

To better understand the effectiveness of our proposed BvSP in predicting implicit information, it is separately evaluated in four different few-shot settings.
The results of these comparisons are depicted in Table \ref{table:implicit information}. Our testing set is divided into two subsets. The explicit subset refers to both the aspect term and opinion term being explicit. Then if any of them is implicit, it is put into the implicit subset. 

We can observe that our method BvSP outperforms baselines under all four few-shot settings, demonstrating superior performance in both explicit and implicit information prediction. In particular, compared to the robust baseline GAS, BvSP achieves a notable enhancement of +8.32\% and +6.34\% in the average F1 score for explicit and implicit information, respectively.

Furthermore, it's worth noting that BvSP's performance improves for both explicit and implicit subsets as the number of shots increases. The F1 score for the implicit subset shows a similar upward trend as for the explicit one. This finding highlights the model’s reliance on ample data to enhance its performance. 
Otherwise, the uneven distribution of explicit and implicit information across quad categories in the FSQP dataset could explain the variations in results between models.

\subsection{Performance of Different Language Models}

    

\begin{table}[!t]
    \centering
    \setlength{\tabcolsep}{1.0mm}{\begin{tabular}{l c c}
    \toprule
    Baseline & One-shot & Five-shot \\

    \midrule
    ChatGPT & 23.56 & 38.97 \\
    Llama-2-70B-Chat & 22.91 & 25.95 \\
    Mixtral-8x7B-Instruct & 17.38 & 23.15 \\
    GPT-4 & 28.50 & 40.09 \\
    \midrule
    BvSP-T5-Base & \textbf{37.99} & \textbf{53.52} \\
    BvSP-Shearing-Llama & 25.19 & 47.86 \\
    \bottomrule
    \end{tabular}}
    \caption{Performance comparison of various language models under one-shot and five-shot settings. F1-score (\%) metrics are reported for each model.}
    \label{table:other_llm_results}
\end{table}

We expand our evaluation to include three additional large language models: Llama-2-70B-Chat, Mixtral-8x7B-Instruct, and GPT-4. We also introduced a trainable LLM, shearing-llama-370M \cite{xia2023sheared}.

As shown in Table \ref{table:other_llm_results}, BvSP-T5-Base consistently outperforms other models in both settings, achieving the highest F1 of 37.99\% in the one-shot and 53.52\% in the five-shot scenarios. This indicates its superiority to generalize from limited examples. 

GPT-4 also shows strong performance, particularly in the five-shot setting, highlighting its robustness with additional context. 
Other models like ChatGPT and Llama-2-70B-Chat demonstrate moderate performance, with notable improvements in the five-shot setting compared to the one-shot. Mixtral-8x7B-Instruct performs lower overall.
However, due to the inference length limitation of LLMs, errors often occur when the validated shot number exceeds five. Thus, we speculate that without additional training, increasing the inference length of LLMs could accommodate more shots and enhance the performance of LLM evaluations.

We also observe that the performance of BvSP-Shearing-Llama is not as good as BvSP-T5-Base. Two possible explanations are: (1) the effect of training parameters, suggesting that we might need to adjust them for Shearing-Llama, and (2) differences in model characteristics. Shearing-Llama is derived from the general model Llama-7b through pruning, focusing more on general tasks like lm-evaluation-harness. In contrast, T5-Base is pre-trained to solve downstream tasks, including sentiment analysis, giving it a potential advantage over Shearing-Llama.

\subsection{Training Time Analysis}

\begin{table}[t!]
    \centering
    \setlength{\tabcolsep}{2.0mm}{
    \begin{tabular}{c c c c}
    \toprule
    \textbf{Models} & One-shot \\
    \midrule
    GAS & 216s \\
    Paraphrase & 201s \\
    DLO & 418s \\ 
    ILO & 436s \\
    \midrule
    BvSP & 477s \\
    \bottomrule
    \end{tabular}}
    \caption{Average running time of each model.}
    \label{table:time}
\end{table}

The average running time of each model is shown in Table \ref{table:time}. We can observe that on five generation-based methods, BvSP consistently causes more training time and consumes a little more time compared with ILO.

\section{Supplementary Materials of $\mathtt{FSQP}$}\label{sec:Supplementary Materials}

For our dataset $\mathtt{FSQP}$, we present the complete set of categories in the training set, development set, and testing set in Table \ref{table:aspect_category}. We further select two comments from each of the ten categories from these sets. The sampled reviews are demonstrated in Table \ref{table:case_for_train_set}, \ref{table:case_for_dev_set}, and \ref{table:case_for_test_set}. The first column displays the review sentence. And the second column shows the extracted quads (aspect terms, aspect category, sentiment polarity, opinion terms).

\begin{table*}[]
\small
    \centering
    \setlength{\tabcolsep}{2.0mm}{\begin{tabular}{l l}
    \toprule
    & \multicolumn{1}{c}{\textbf{Aspect category}} \\
    \midrule
    \multirow{7}{*}{\textbf{Train}} & room\_overall, room\_smoke, service\_staff, service, price, salon, food\_dessert, food, experience, hotel, \\
    & service\_staff\_doctor, building\_hall, food\_meat\_chicken, procedure\_beauty\_nails, food\_meat, \\
    & building\_elevator, location, decor, room\_bathroom, room\_equipment, room\_interior, \\
    & procedure\_relax\_massage, procedure\_beauty\_nails\_pedi, procedure\_beauty\_face, food\_salad, \\
    & service\_staff\_front-desk, food\_bread, procedure\_beauty\_nails\_mani, procedure\_beauty\_hair, \\
    & food\_meat\_beef, drinks, parking, service\_staff\_master, procedure\_beauty\_wax, service\_staff\_owner, \\
    & food\_meat\_pork, food\_mealtype\_start, cleanliness, food\_meat\_rib, food\_portion, salon\_equipment, \\
    & room\_bed, room\_bedroom, food\_fruit, internet \\
    \midrule
    \multirow{3}{*}{\textbf{Dev}} & food\_side\_potato, restaurant, food\_meat\_steak, food\_side\_vegetables, food\_meat\_burger, food\_cheese, \\
    & food\_side\_pasta, food\_mealtype\_main, procedure\_relax\_train, food\_sushi, salon\_additional, food\_seafood, \\
    & salon\_interior, salon\_interior\_bath, salon\_interior\_room \\
    \midrule
    \multirow{4}{*}{\textbf{Test}} & food\_eggs, food\_soup, procedure\_beauty\_barber, service\_management, procedure\_relax\_spa, \\
    & salon\_atmosphere, food\_mealtype\_breakfast, drinks\_alcohol\_light, entertainment\_atmosphere, sport\_pool, \\
    & restaurant\_atmosphere, food\_selection, building, occasion, drinks\_alcohol\_beer, service\_staff\_waiter, \\
    & drinks\_non-alcohol, food\_mealtype\_dinner, drinks\_alcohol\_wine, service\_wait \\
    \bottomrule
    \end{tabular}}
    \caption{The aspect categories contained in the training set, development set, and testing set are shown.}
    \label{table:aspect_category}
\end{table*}

\begin{table*}[]
\small
    \centering
    \setlength{\tabcolsep}{2.0mm}{\begin{tabular}{l l}
    \toprule
    \multicolumn{2}{l}{\textbf{room\_overall}} \\
    \midrule
    If you are the type to stay in your room a lot then this is the place for you . & (room, room\_overall, positive, NULL) \\
    \midrule
    \multirow{2}{*}{We were very disappointed with our three day stay at red rock in room 16143 .} & (room, room\_overall, negative, \\
    & very disappointed) \\
    \bottomrule
    \multicolumn{2}{l}{\textbf{service}} \\
    \midrule
    \multirow{2}{*}{They are very professional and great customer service .} & (customer service, service, positive, \\
    & professional and great) \\
    \midrule
    \multirow{1}{*}{We would have enjoyed more if not for the unprofessional wait staff .} & (staff, service, negative, unprofessional) \\
    \bottomrule
    \multicolumn{2}{l}{\textbf{price}} \\
    \midrule
    \multirow{1}{*}{The prices are also reasonable .} & (prices, price, positive, reasonable) \\
    \midrule
    \multirow{1}{*}{I did n't bother going back it seems like a waste of time and money ...} & (money, price, negative, waste) \\
    \bottomrule
    \multicolumn{2}{l}{\textbf{food}} \\
    \midrule
    \multirow{1}{*}{I had the vodka penne and it was delicious !} & (vodka penne, food, positive, delicious) \\
    \midrule
    \multirow{1}{*}{I have never had such disgusting chinese food then this .} & (chinese food, food, negative, disgusting) \\
    \bottomrule
    \multicolumn{2}{l}{\textbf{decor}} \\
    \midrule
    \multirow{1}{*}{I 'm a huge phillip stark fan and i thought the decor was beautiful .} & (decor, decor, positive, beautiful) \\
    \midrule
    \multirow{1}{*}{Currently staying here , carpet is gross .} & (carpet, decor, negative, gross) \\
    \bottomrule
    \multicolumn{2}{l}{\textbf{internet}} \\
    \midrule
    \multirow{1}{*}{Wifi works best on odd room numbers .} & (wifi, internet, positive, works best) \\
    \midrule
    \multirow{2}{*}{The sign on the wall gives you a wifi password to use , but it doesn't work .} & (wifi password, internet, negative, \\
    & doesn't work) \\
    \bottomrule
    \multicolumn{2}{l}{\textbf{cleanliness}} \\
    \midrule
    \multirow{2}{*}{First the cleanliness of the room was substandard .} & (cleanliness, cleanliness, negative, \\
    & substandard) \\
    \midrule
    \multirow{2}{*}{I was disappointed by the cleanliness of the room .} & (cleanliness, cleanliness, negative,\\
    & disappointed) \\
    \bottomrule
    \multicolumn{2}{l}{\textbf{drinks}} \\
    \midrule
    \multirow{1}{*}{The drinks were also delicious !} & (drinks, drinks, positive, delicious) \\
    \midrule
    \multirow{1}{*}{Mmmm ... the drinks are n't that good!} & (drinks, drinks, negative, n't that good) \\
    \bottomrule
    \multicolumn{2}{l}{\textbf{location}} \\
    \midrule
    \multirow{2}{*}{This is one of my favorite sephora locations .} & (sephora locations, location, positive, \\
    & favorite) \\
    \midrule
    \multirow{1}{*}{The only negative about this place is the location .} & (location, location, negative, negative) \\
    \bottomrule
    \multicolumn{2}{l}{\textbf{hotel}} \\
    \midrule
    \multirow{1}{*}{We loved this hotel !} & (hotel, hotel, positive, loved) \\
    \midrule
    \multirow{1}{*}{I will not be staying in this hotel again ...} & (hotel, hotel, negative, not be staying in) \\
    \bottomrule
    \end{tabular}}
    \caption{Sampled reviews from the training set of $\mathtt{FSQP}$.}
    \label{table:case_for_train_set}
\end{table*}

\begin{table*}[]
\small
    \centering
    \setlength{\tabcolsep}{2.0mm}{\begin{tabular}{l l}
    \toprule
    \multicolumn{1}{l}{\textbf{food\_side\_potato}} \\
    \midrule
    \multirow{2}{*}{We also ordered potatoes au gratin which were amazing ! ! !} & (potatoes au gratin, food\_side\_potato, positive, \\
    & amazing) \\
    \midrule
    \multirow{2}{*}{The potato skins were tiny and limp .} & (potato skins, food\_side\_potato, negative, \\
    & tiny and limp) \\
    \bottomrule
    \multicolumn{1}{l}{\textbf{restaurant}} \\
    \midrule
    \multirow{1}{*}{Acoustically , the restaurant was distracting .} & (restaurant, restaurant, negative, distracting) \\
    \midrule
    \multirow{1}{*}{Finally a reliable chinese restaurant !} & (chinese restaurant, restaurant, positive, reliable) \\
    \bottomrule
    \multicolumn{1}{l}{\textbf{food\_meat\_steak}} \\
    \midrule
    \multirow{1}{*}{The cut they suggested , was along the best steak i 've ever had .} & (steak, food\_meat\_steak, positive, best) \\
    \midrule
    \multirow{1}{*}{The steak was so rough and disgusting i actually cut it up} & \multirow{1}{*}{(steak, food\_meat\_steak, negative, } \\
    and fed it to my steaks . & rough and disgusting)\\
    \bottomrule
    \multicolumn{1}{l}{\textbf{food\_side\_vegetables}} \\
    \midrule
    \multirow{1}{*}{The veggies with the sauces excellent !} & (veggies, food\_side\_vegetables, positive, excellent) \\
    \midrule
    \multirow{2}{*}{I was a bit disappointed by the fried pickles} & (fried pickles, food\_side\_vegetables, negative, \\ 
    & disappointed) \\
    \bottomrule
    \multicolumn{1}{l}{\textbf{food\_meat\_burger}} \\
    \midrule
    \multirow{2}{*}{I ordered my wineburger it did not disappoint .} & (wineburger, food\_meat\_burger, positive, \\
    & not disappoint) \\
    \midrule
    \multirow{2}{*}{Burgers are very under cooked !} & (burgers, food\_meat\_burger, negative, \\
    & very under cooked) \\
    \bottomrule
    \multicolumn{1}{l}{\textbf{food\_cheese}} \\
    \midrule
    \multirow{1}{*}{The cotija cheese was yummy .} & (cotija cheese, food\_cheese, positive, yummy) \\
    \midrule
    \multirow{1}{*}{The cheese had no flavor .} & (cheese, food\_cheese, negative, no flavor) \\
    \bottomrule
    \multicolumn{1}{l}{\textbf{procedure\_relax\_train}} \\
    \midrule
    \multirow{1}{*}{This gym has everything i need ... if only i could step it up} & \multirow{2}{*}{(gym, procedure\_relax\_train, positive, NULL)} \\
     so that i actually see results ! & \\
    \midrule
    \multirow{1}{*}{The gym ... ... was crowded .} & (gym, procedure\_relax\_train, negative,  crowded) \\
    \bottomrule
    \multicolumn{1}{l}{\textbf{food\_sushi}} \\
    \midrule
    \multirow{1}{*}{The sushi was really good !} & (sushi, food\_sushi, positive, good) \\
    \midrule
    \multirow{1}{*}{Cons - sushi pieces were smaller than i expected .} & (sushi pieces, food\_sushi, negative, smaller) \\
    \bottomrule
    \multicolumn{1}{l}{\textbf{food\_seafood}} \\
    \midrule
    \multirow{2}{*}{The seafood fradiavolo was delicious .} & (seafood fradiavolo, food\_seafood, positive, \\
    & delicious) \\
    \midrule
    \multirow{1}{*}{If you are expecting legal seafood do n't go here .} & (seafood, food\_seafood, negative, NULL) \\
    \bottomrule
    \multicolumn{1}{l}{\textbf{salon\_interior}} \\
    \midrule
    \multirow{1}{*}{The bed is super comfortable  .} & (bed, salon\_interior, positive, comfortable) \\
    \midrule
    \multirow{1}{*}{Seriously , it was the worst sofa bed in the world .} & (sofa bed, salon\_interior, negative, worst) \\
    \bottomrule
    \end{tabular}}
    \caption{Sampled reviews from the development set of $\mathtt{FSQP}$.}
    \label{table:case_for_dev_set}
\end{table*}

\begin{table*}[]
\small
    \centering
    \setlength{\tabcolsep}{2.0mm}{\begin{tabular}{l l}
    \toprule
    \multicolumn{2}{l}{\textbf{food\_eggs}} \\
    \midrule
    The eggs benedict were delicious ! & (eggs benedict, food\_eggs, positive, delicious) \\
    \midrule
    \multirow{1}{*}{The egg rolls were slighty burnt .} & (egg rolls, food\_eggs, negative, slighty burnt) \\
    \bottomrule
    \multicolumn{2}{l}{\textbf{food\_soup}} \\
    \midrule
    \multirow{1}{*}{We had miso soup to start , was great !} & (miso soup, food\_soup, positive, great) \\
    \midrule
    \multirow{1}{*}{Avoid the french onion soup .} & (french onion soup, food\_soup, negative, avoid) \\
    \bottomrule
    \multicolumn{2}{l}{\textbf{service\_management}} \\
    \midrule
    \multirow{1}{*}{Manager came by a few times to be sure we were} & \multirow{2}{*}{(manager, service\_management, positive, NULL)} \\
    satisfied . & \\
    \midrule
    \multirow{1}{*}{Manager went out of her way to apologize but} & \multirow{2}{*}{(manager, service\_management, negative, disappointed)} \\
    by then we were very very disappointed . & \\
    \bottomrule
    \multicolumn{2}{l}{\textbf{procedure\_relax\_spa}} \\
    \midrule
    \multirow{1}{*}{My friend and i had a perfect spa day package there .} & (spa day, procedure\_relax\_spa, positive, perfect) \\
    \midrule
    \multirow{1}{*}{Do n't waste your time at any other spa on the strip .} & (spa, procedure\_relax\_spa, negative, NULL) \\
    \bottomrule
    \multicolumn{2}{l}{\textbf{drinks\_alcohol\_light}} \\
    \midrule
    \multirow{1}{*}{Their signature pineapple martini was to die for !} & (pineapple martini, drinks\_alcohol\_light, positive, to die for) \\
    \midrule
    \multirow{1}{*}{Martini was short and missing several sips .} & (martini, drinks\_alcohol\_light, negative, short) \\
    \bottomrule
    \multicolumn{2}{l}{\textbf{entertainment\_atmosphere}} \\
    \midrule
    \multirow{1}{*}{Also the noise level was a little high .} & (noise level, entertainment\_atmosphere, negative, little high) \\
    \midrule
    \multirow{1}{*}{Very nice people and great atmosphere !} & (atmosphere, entertainment\_atmosphere, positive, great) \\
    \bottomrule
    \multicolumn{2}{l}{\textbf{sport\_pool}} \\
    \midrule
    \multirow{1}{*}{Pools are great in the summer .} & (pools, sport\_pool, positive, great) \\
    \midrule
    \multirow{1}{*}{The pool had trash in it .} & (pool, sport\_pool, negative, had trash in) \\
    \bottomrule
    \multicolumn{2}{l}{\textbf{building}} \\
    \midrule
    \multirow{1}{*}{It looks like an old mayan temple building in} & \multirow{2}{*}{(mayan temple building, building, negative, old)} \\
    need of updates . & \\
    \midrule
    \multirow{1}{*}{But it is a beautiful building .} & (building, building, positive, beautiful) \\
    \bottomrule
    \multicolumn{2}{l}{\textbf{food\_selection}} \\
    \midrule
    \multirow{1}{*}{Anything you want thats on the menu .} & (menu, food\_selection, positive, NULL) \\
    \midrule
    \multirow{1}{*}{But the menu is really very limited .} & (menu, food\_selection, negative, limited) \\
    \bottomrule
    \multicolumn{2}{l}{\textbf{service\_wait}} \\
    \midrule
    \multirow{1}{*}{While you are waiting you can sit comfortably} & \multirow{2}{*}{(waiting, service\_wait, positive, comfortably)} \\
    on one of the couches . & \\
    \midrule
    \multirow{1}{*}{So there i sat waiting even longer ! ! ! } & (waiting, service\_wait, negative, even longer) \\
    \bottomrule
    \end{tabular}}
    \caption{Sampled reviews from the testing set of $\mathtt{FSQP}$.}
    \label{table:case_for_test_set}
\end{table*}

\end{document}